\pdfoutput=1
\documentclass[11pt]{article}

\usepackage[final]{acl}

\usepackage{natbib}
\usepackage{xcolor}
\usepackage{hyperref}
\usepackage{times}
\usepackage{latexsym}
\usepackage{multirow}
\usepackage{multicol}
\usepackage{graphicx}
\usepackage{siunitx}
\usepackage[most]{tcolorbox}

\usepackage[textsize=scriptsize]{todonotes}
\usepackage{inconsolata}
\usepackage[T1]{fontenc}
\usepackage[utf8]{inputenc}

\usepackage{microtype}

\usepackage{booktabs} % To thicken table lines
\usepackage{amsmath}
\usepackage{mathtools}
\usepackage{amssymb}
\usepackage{lineno}
\usepackage{enumitem}

\usepackage[noend]{algpseudocode}
\usepackage{algorithm}

\usepackage{array} % Required for custom column formatting

\definecolor{todoblue}{RGB}{33,150,243}

\definecolor{pink-color}{RGB}{255,1,97} 

\definecolor{red-color}{RGB}{255,111,105} 
\definecolor{dark-red-color}{RGB}{174,0,1} 
\definecolor{light-red-color}{RGB}{255,152,150} 

\definecolor{green-color}{RGB}{136,216,176} 
\definecolor{dark-green-color}{RGB}{49,163,84} 
\definecolor{light-green-color}{RGB}{136,216,176} 
\definecolor{sea-green-color}{RGB}{95,158,160} 

\definecolor{dark-blue-color}{RGB}{49,130,189} 
\definecolor{light-blue-color}{RGB}{158,202,225} 

\definecolor{grey-color}{RGB}{167,173,186}
\definecolor{dark-grey-color}{RGB}{79,91,102}
\definecolor{gold-color}{RGB}{247,205,15} 

\newcommand{\OPRO}{\mbox{\textsc{OPRO}}}
\newcommand{\GPO}{\mbox{\textsc{GPO}}}

\newcommand{\GEPA}{\mbox{\textsc{GEPA}}}

\newcommand{\MOPO}{\mbox{\textsc{MOPO}}}
\newcommand{\ProTeGi}{\mbox{\textsc{ProTeGi}}}
\newcommand{\EvoPrompt}{\mbox{\textsc{EvoPrompt}}}

\newcommand{\SummEval}{\mbox{\textsc{SummEval}}}

\newcommand{\BRIGHTER}{\mbox{\textsc{BRIGHTER}}}

\newcommand{\insection}[2][.]{{\setlength{\parskip}{6pt} \noindent\textbf{#2#1}}}

\newcommand{\ignore}[1]{}
\newcommand{\ig}[1]{}

\newcolumntype{h}{>{\setbox0=\hbox\bgroup}c<{\egroup}@{}}
\newcolumntype{C}[1]{>{\centering\arraybackslash}m{#1}}
\newcolumntype{R}[1]{>{\raggedleft\arraybackslash}p{#1}}
\newcolumntype{L}[1]{>{\raggedright\arraybackslash}p{#1}}

\newcommand{\urlsmall}[1]{{\scriptsize{\url{\detokenize{#1}}}}}

\newtcolorbox[list inside=prompt,auto counter,number within=section]{prompt}[1][]{
    colbacktitle=black!60,
    coltitle=white,
    fontupper=\footnotesize,
    boxsep=5pt,
    left=0pt,
    right=0pt,
    top=0pt,
    bottom=0pt,
    boxrule=1pt,
    #1,
}

\title{When Gradients Collide: Failure Modes of Multi-Objective Prompt Optimization for LLM Judges}

\author{
\hspace{1.0cm} Parth Darshan\textsuperscript{$\diamondsuit$}\thanks{Equal contribution. $^\dagger$Research lead and corresponding author. Author was employed at Amazon when this work was conducted, but the work was performed independently and did not use Amazon data, resources or confidential information.}  \hspace{2.25cm} Abhishek Divekar\textsuperscript{$\spadesuit \dagger$*}\\
\textsuperscript{$\diamondsuit$}IIT Jodhpur \hspace{3.0cm}
\hspace{0.5cm}\textsuperscript{$\spadesuit$}Amazon \\
\hspace{0.5cm} \texttt{b22cs040@iitj.ac.in} \hspace{1.5cm} \texttt{adivekar@amazon.com} \\
}

\begin{document}
\maketitle

\begin{abstract}
Customizing an LLM judge to a specific problem or domain often involves optimizing its prompt across multiple evaluation criteria simultaneously.
Textual gradient methods automate this for a single judge criterion, however they produce natural-language critiques, not numerical vectors.
Thus, the conflict-resolution toolkit of multi-task learning (PCGrad, MGDA) does not apply to this multi-objective textual gradient setting.    
We extend TextGrad to the multi-objective setting and test four decomposition modes of textual gradient optimizers by varying how much cross-objective information the loss, gradient and optimizer LLMs share.
We find the gradient's task-focus drops by 59\% (9.0 to 3.7 out of 10) when the gradient LLM must provide feedback on multiple criteria jointly.
Separately, we observe that naively combining single-objective optimized instructions into a single prompt degrades Spearman~$\rho$ from 0.305 to 0.220 ($-0.085$).
These results identify two separable failure modes: optimization-time \emph{gradient dilution} and inference-time \emph{instruction interference}, which together constrain the design space for multi-objective judge optimization using textual feedback.\footnote{Code, data and prompts available at \url{https://github.com/adivekar-utexas/when-gradients-collide}}
\end{abstract}
    
\section{Introduction}
\label{sec:introduction}

\begin{figure*}[t]
\centering
\includegraphics[width=0.9\textwidth]{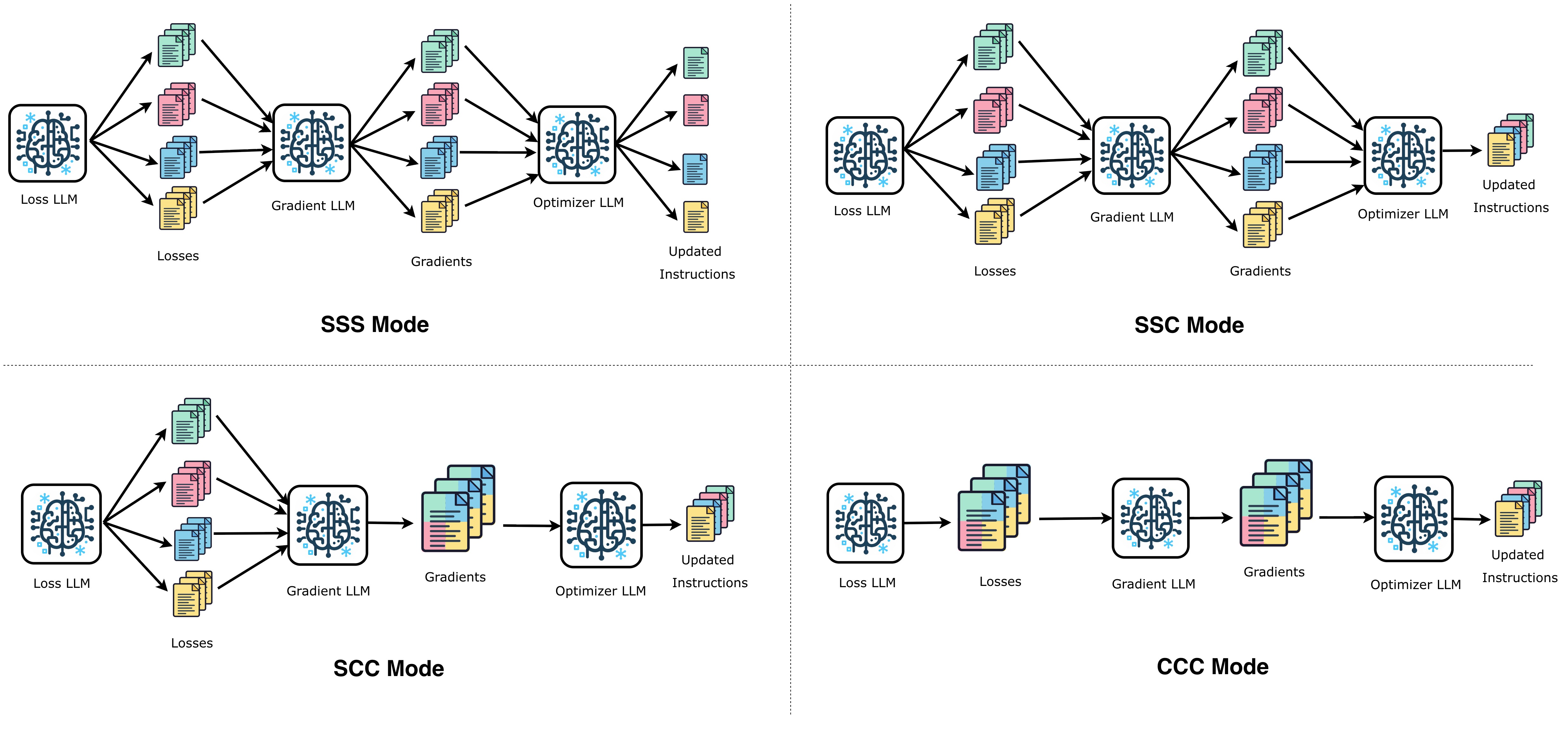}
\caption{Overview of the optimization pipeline. Each step consists of four stages: (1)~the task model predicts scores, (2)~the loss LLM critiques predictions against ground truth, (3)~the gradient LLM converts critiques into instruction edits, and (4)~the optimizer LLM rewrites the prompt. The decomposition mode determines whether each stage operates per-task (Separate) or over all tasks jointly (Combined). Only the per-task instruction text is updated; the prompt skeleton and output format remain frozen throughout optimization.}
\label{fig:overview}
\end{figure*}

\vspace{-0.25em}
Modern judges, whether human or LLMs, are used to evaluate text along multiple quality dimensions at once.
\SummEval{} \citep{fabbri-etal-2021-summeval} scores summaries on four dimensions simultaneously; MT-Bench \citep{zheng2023judging} spans eight categories from coding to roleplay.
\citet{kim2024prometheus2} developed Prometheus~2 specifically because existing evaluators ``do not possess the ability to evaluate based on \emph{custom evaluation criteria}''.
Prompt optimization methods like TextGrad \citep{yuksekgonul2025textgrad}, \OPRO{} \citep{yang2024opro}, and \GEPA{} \citep{agrawal2026gepa} can improve prompts for LLM judges automatically, but they optimize a single objective.
Whether they extend to multi-objective judges, where one prompt must score several evaluation dimensions at once, remains less studied.

The core challenge is structural. 
When multi-task deep learning models encounter conflicting task gradients, methods like PCGrad \citep{yu2020pcgrad} and MGDA \citep{sener2018moo} resolve conflicts via projection or constrained optimization.
The conflict-resolution tools of numerical multi-task learning do not apply directly, because textual gradients lack the vector-space structure these methods require.
As textual gradients are natural-language strings, they do not have equivalent concepts of magnitude, inner products, or linear subspaces.
The instruction ``Make the coherence rubric more specific'' cannot be numerically projected against ``simplify the fluency criteria''.

Prior work addresses related problems but not this intersection.
For example, intra-task feedback quality has been studied: recently \citet{chu2026caro} identify \emph{rule dilution}, when heterogeneous error modes are aggregated within a single optimization step, and \citet{melcer2025textgrad} show that the gradient analogy does not accurately explain why Automatic Prompt Optimization (APO) methods succeed.
Multi-objective prompt optimization also exists: \citet{jafari-etal-2024-morl} find that MGDA underperforms volume-based alternatives for discrete prompt tokens. Multi-task judge \emph{training} is studied extensively \citep{whitehouse2026j1, yang2026fairjudge, wang2025djpo}, but these methods update weights, not prompts.
To our knowledge, no prior work studies multi-objective judge \emph{prompt} optimization using textual gradients.

% Multi-agent coordination has also been explored: \citet{han2025mapgd} coordinate dimension-specialized prompt agents via a semantic gradient coordinator, but optimize a single objective.

% Concretely, we investigate three research questions: (i) does multi-task \TextGrad{} improve judge prompts over the initial prompt? (ii) What process-level mechanisms explain failures to optimize? (iii) Does inference-time instruction interference compound optimization-time failures?

We propose and test four plausible decomposition modes of textual gradient optimization for multi-objective judges: \textsc{SSS}, \textsc{SSC}, \textsc{SCC}, \textsc{CCC}, where the naming encodes whether each pipeline stage (loss, gradient, optimizer) processes tasks separately~(S) or combined~(C).
We also introduce two process-level diagnostics: \emph{gradient specificity} (how targeted each gradient is to a single task) and \emph{feedback adherence} (whether the optimizer follows the gradient). We evaluate on \SummEval{} \citep{fabbri-etal-2021-summeval} with four criteria, using two validation gate settings (MAE gating and no gating).

Our results reveal a consistent structure in how and why multi-criteria prompt optimization stagnates.
In 6 of 10 configurations with Qwen3 judges, optimization never exceeds the initial generic prompt (\S\ref{sec:results}, Table~\ref{tab:main-results}); only \textsc{Single-Task} with \texttt{val=mae} improves ($+0.031$ Spearman at step~2). Our diagnostics localize this effect: gradient specificity drops by 59\% ($9.0$ to $3.7$) when the gradient LLM processes all tasks jointly \S\ref{sec:gradient-specificity}, Table~\ref{tab:specificity-pertask}). Feedback adherence remains high (7.8 to 8.8), indicating that the optimizer incorporates the gradient's suggestions regardless of their specificity.
A separate oracle experiment shows that even independently optimal per-task instructions \emph{degrade} from 0.305 to 0.220 ($-0.085$ Spearman) when combined into one prompt (\S\ref{sec:instruction-interference}, Table~\ref{tab:cherrypick}). These results identify two separable failure modes: optimization-time \emph{gradient dilution} and inference-time \emph{instruction interference}.

Our contributions are as follows: 
(1) we present an empirical study of multi-criteria textual gradient optimization for LLM judges across four decomposition modes (\S\ref{sec:setup}, \S\ref{sec:results}, Table~\ref{tab:main-results});
(2) we propose two process-level diagnostics (\emph{gradient specificity} and \emph{feedback adherence}) that localize issues in optimization to gradient quality, not optimizer compliance (\S\ref{sec:gradient-specificity}, Figure~\ref{fig:specificity}, Table~\ref{tab:adherence}).
(3) we conduct an oracle-instruction experiment which isolates inference-time \emph{instruction interference} as a failure mode distinct from optimization-time dilution (\S\ref{sec:instruction-interference}, Table~\ref{tab:cherrypick}).
\section{Related Work}
\label{sec:related}
\vspace{-1em}

\insection{Textual Gradient Methods}
Early methods in Prompt optimization used scalar signals: \OPRO{}~\citep{yang2024opro} rewrites prompts by conditioning on (prompt, score) histories, and APE~\citep{zhou2023ape} generates candidates from demonstrations and selects the highest-scoring variant.
\ProTeGi{}~\citep{pryzant-etal-2023-automatic} replaced scalar signals with \textit{textual gradients}: natural-language critiques that guide prompt rewrites.
TextGrad~\citep{yuksekgonul2025textgrad} extended this to multi-component computation graphs, propagating critiques through LLM pipelines in analogy to backpropagation.

Subsequent work has questioned the gradient analogy.
\GPO{}~\citep{tang2025gpo} decomposed textual optimization into two factors (update direction and update method), drawing analogies to gradient, momentum, and learning rate, but found that adding a reflection step hurts performance.
\citet{melcer2025textgrad} showed empirically that the gradient decomposition (the chain-rule structure central to the analogy) does not consistently improve prompt optimization, and that the gradient metaphor does not accurately explain why APO methods succeed. \GEPA{}~\citep{agrawal2026gepa} evolves language programs via reflective mutation with Pareto selection.
To our knowledge, all methods above optimize a single objective and none provide a mechanism to observe or control how per-task feedback interacts during optimization.

\insection{Multi-Objective Prompt Optimization}
Multi-objective prompt optimization is a nascent area with two distinct approaches.
Population-based methods maintain a Pareto front of candidate prompts: \MOPO{}~\citep{menchaca-resendiz-klinger-2025-mopo} applies NSGA-II~\citep{deb2002nsga2} with LLM-based mutation to affective text generation, and ParetoPrompt~\citep{zhao2025paretoprompt} decomposes objectives into scalarized subproblems. MORL-Prompt~\citep{jafari-etal-2024-morl} adapts multi-objective reinforcement learning to discrete prompt tokens but found that MGDA~\citep{sener2018moo} underperforms the simpler product-of-rewards at balancing competing objectives.
Evolutionary approaches include \EvoPrompt{}~\citep{guo2024connecting} for single-objective and \citet{baumann2024emoprompts} for multi-objective sentiment balancing.
These methods operate at the \textit{candidate-selection} level: they choose which prompts to keep from a population.
To our knowledge, none of the above studies how per-task feedback interacts \textit{within} a single textual gradient trajectory, which is the setting we investigate.

\insection{LLM-as-a-Judge Evaluation}
MT-Bench and Chatbot Arena~\citep{zheng2023judging} established the paradigm of LLM-based evaluation where human annotation is expensive. Subsequent works use a fixed prompt to improve judge quality through tuning weights on rubric-grounded data \citep{kim2024prometheus2}, reinforcement learning \citep{whitehouse2026j1}, debiasing \citep{yang2026fairjudge}, and preference optimization \citep{wang2025djpo}.
A parallel line of work optimizes the evaluation prompt itself.
CARO~\citep{chu2026caro} identifies \textit{rule dilution} when heterogeneous error modes are aggregated into a single optimization step. % within an educational grading rubric.
RRD~\citep{shen2026rrd} recursively decomposes rubrics to improve coverage and remove redundancy.
MPO~\citep{sharma2026mpo} applies section-local textual gradients to individual prompt components (role, context, constraints) independently.
MAPGD~\citep{han2025mapgd} coordinates multiple gradient agents, using semantic similarity to resolve conflicting edits. All these operate on a single evaluation objective, or decompose along axes other than evaluation criteria.
None jointly optimizes a judge prompt across multiple criteria while preserving per-task gradient observability (i.e., the ability to trace how each criterion's feedback shaped each rewrite), the setting we study.

\begin{table*}[t!]
\centering
\small
\begin{tabular}{l cccc cccc}
\toprule
& \multicolumn{4}{c}{\textbf{MAE validation}} & \multicolumn{4}{c}{\textbf{No validation}} \\
\cmidrule(lr){2-5} \cmidrule(lr){6-9}
\textbf{Mode} & \textbf{Initial} & \textbf{Best (step)} & \textbf{$\Delta$} & \textbf{HVI} & \textbf{Initial} & \textbf{Best (step)} & \textbf{$\Delta$} & \textbf{HVI} \\
\midrule
\textsc{Single-Task}  & 0.274 & \textbf{0.305} (2) & \textbf{+0.031} & ---   & 0.269 & 0.284 (5) & \textbf{+0.015} & ---   \\
[0.5ex]
\textsc{SSS}     & 0.284 & 0.284 (0)       & +0.000          & 2.749 & 0.283 & 0.283 (0)       & +0.000          & 2.867 \\
\textsc{SSC}     & 0.289 & 0.289 (0)       & +0.000          & 2.832 & 0.283 & \textbf{0.291} (2)  & +0.007          & 2.845 \\
\textsc{SCC}     & 0.282 & 0.282 (0)       & +0.000          & 2.801 & 0.282 & 0.282 (0)       & +0.000          & 2.779 \\
\textsc{CCC}     & 0.285 & 0.296 (9)       & +0.012          & \textbf{2.900} & 0.287 & 0.287 (0)       & +0.000          & \textbf{2.983} \\
\bottomrule
\end{tabular}
\caption{Main results on \SummEval{} using Qwen3. We measure the task-averaged Spearman correlation (mean over $N\!=\!3$ runs). $\Delta$ is the improvement from the initial prompt to the best step. HVI is the hypervolume indicator at step~12 (higher $=$ more Pareto-diverse prompts). We evaluate four decomposition modes (SSS through CCC) plus a single-task optimization baseline (\textsc{Single-Task}), covering the full spectrum from separate to joint optimization. In 6 of 10 configurations, the initial generic prompt (``\textit{Rate from 1 to 5}'') is never exceeded.}
\label{tab:main-results}
\end{table*}

\section{Method: Multi-objective TextGrad}
\label{sec:method}

\vspace{-1em}
\insection{TextGrad pipeline}
We implement a 4-stage optimization loop based on TextGrad \cite{yuksekgonul2025textgrad}.
At each step, a \textbf{task LLM} predicts objective-scores for a minibatch of examples using the current prompt.
A \textbf{loss LLM} critiques the predictions by comparing them to ground-truth annotations and produces a natural-language ``loss'' for each example (called a \textit{reflection} by some authors).
A \textbf{gradient LLM} aggregates the per-example losses into a ``textual gradient'': a structured set of instruction-level edit suggestions. We restrict gradients to 3 paragraphs.
An \textbf{optimizer LLM} rewrites the prompt based on the gradient.
Only the per-objective instructions are modified; the prompt skeleton (role preamble, output format, few-shot examples) is frozen throughout (see Appendix~\ref{sec:appendix-prompts} for the full initial prompt and an example of optimized instructions).
This design isolates the effect of objective-level instruction modifications from possible confounds introduced by structural prompt changes.

\insection{Decomposition modes}
We extend TextGrad to the multi-objective setting stagewise. As shown in Figure~\ref{fig:overview}, we parameterize the multi-objective interaction via a 3-letter \textit{decomposition code}. Each letter denotes whether a stage operates in \textbf{S}eparate (per-task) or \textbf{C}ombined (all-tasks-joint) mode, applied to the loss, gradient, and optimizer stages respectively.

Four modes span the design space.
\textbf{SSS} operates all three stages independently per task.
\textbf{SSC} computes loss and gradient per task, but the optimizer receives all four gradients simultaneously and rewrites the prompt in a single call.
\textbf{SCC} computes only the loss per task; the gradient LLM receives critiques from all four tasks and produces a unified set of edits.
\textbf{CCC} operates all stages jointly.
We also include a \textbf{Single-Task} baseline in which each objective is optimized in a completely independent run, and evaluated through its own forward pass; i.e., the reported Spearman is the average of four per-criterion evaluations, each using a prompt specialized to a single objective.

The key architectural boundary lies between SSC and SCC. In SSC, the gradient LLM sees one task at a time. In SCC, it must reconcile feedback from all four tasks into a coherent edit plan.

\insection{Validation Gating}
We evaluate two validation strategies.
Under \textbf{val=mae}, a candidate prompt is accepted only if its mean absolute error on a held-out validation set does not exceed that of the current prompt. This acts as a monotonic filter that prevents prompt-regression.
Under no validation gate, every candidate is accepted unconditionally; we can observe the optimization trajectory without gating.
For each configuration, we run 3 independent trials with different random seeds, each for 12 optimization steps.
We report the mean and standard deviation of task-averaged Spearman $\bar{\rho}$ across the 3 runs.

\section{Experimental Setup}
\label{sec:setup}

\vspace{-1em}
\insection{Datasets}
We evaluate on \SummEval{}~\citep{fabbri-etal-2021-summeval}, a summarization meta-evaluation benchmark.
The original dataset contains source news articles, each paired with summaries from 16 different summarization systems, with expert annotations from multiple annotators per pair. 
We randomly subsample this into 160 pairs for training (used as optimization batches) and 480 pairs for held-out evaluation.
Each pair is scored on four dimensions (fluency, relevance, coherence, consistency) on a 1-5 scale.

These four dimensions are the tasks in our multi-task optimization setting: the judge prompt must produce accurate scores across all four simultaneously. We report Spearman rank correlation ($\rho$) between predicted and human scores as the primary metric, following prior work on LLM-based evaluation \citep{Liu2023GEvalNE}.
Unless stated otherwise, all reported results are \textit{task-averaged} Spearman, the arithmetic mean of per-task $\rho$ values across the four dimensions.

\insection{Models}
Our main results are evaluated on the Qwen3 family \citep{yang2025qwen3technicalreport} using Qwen3-8B as the task LLM and Qwen3-235B-A22B as others. 
We use a higher optimizer temperature ($T=0.7$) to encourage diverse rewrites, and a lower loss and gradient temperatures ($T=0.3$) promote consistent critiques.
For the task LLM we set $T=1.0$ to allow resampling in case of JSON formatting errors. In Appendix~\ref{sec:appendix-deepseek}, we show results on DeepSeek v4 Flash and Pro \citep{deepseekai2026deepseekv4}.
\begin{figure*}[t!]
    \centering
    \includegraphics[width=0.95\textwidth]{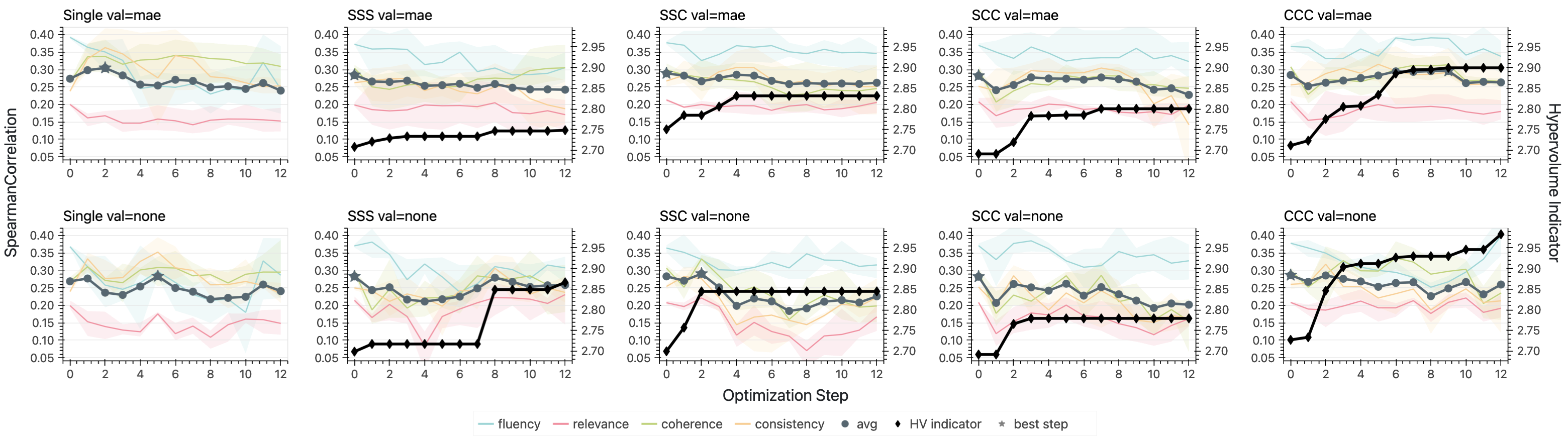}
    \caption{Per-task Spearman $\rho$ for each optimization steps on \SummEval{} with Qwen3. We average over $N\!=\!3$ runs (shaded bands show min to max). Each column shows one of the five decomposition modes. On the top row we apply validation-MAE to gate prompts at each step, while bottom row has no gating. Gray line indicates task-averaged $\rho$; stars mark best step. Black diamonds (right axis) denote the hypervolume indicator for the prompt-candidates accumulated over steps. Without validation gating, multi-task modes degrade, and with it, they plateau.}
    \label{fig:trajectories}
\end{figure*}

\section{Results}
\label{sec:results}

Table~\ref{tab:main-results} reports task-averaged Spearman~$\rho$ for each decomposition mode and validation configuration, averaged over $N\!=\!3$ runs.
In 6 of 10 configurations, the best prompt is the initial generic prompt (``\textit{Rate from 1 to 5}''): optimization either fails to improve or actively degrades performance.
The only multi-task mode that improves is CCC with MAE validation and SSC without validation, which achieve modest gain over 12 steps.
Only the single-task baseline with MAE validation achieves meaningful improvement: $+0.031$ Spearman at step~2 (Table~\ref{tab:main-results}, \textbf{bold}).
This confirms that the TextGrad pipeline can improve individual-task prompts when gradient signal is not contaminated by cross-task information.

We plot optimization trajectories in Figure~\ref{fig:trajectories}, revealing the dynamics behind these aggregates.
Without a validation gate (\texttt{val=none}, bottom row), SSC drops from 0.283 to 0.184 by step~7; SCC shows comparable degradation.
The optimizer proposes changes that improve the training-batch loss but harm held-out generalization. MAE validation partially mitigates this by rejecting spurious updates.
With the validation gate active (top row), trajectories flatten rather than decline: the gate prevents catastrophic degradation but cannot compensate for uninformative gradients.

Performance degrades roughly as Single-Task $>$ SSS,  SSC, SCC, CCC without validation gating. This is consistent with the hypothesis that increasing cross-task coupling amplifies gradient dilution.
The pattern is non-monotonic under MAE validation, however: CCC slightly outperforms SSC. Full coupling may occasionally produce complementary gradients that survive the validation gate.
The process-level diagnostics in \S\ref{sec:analysis} disentangle these effects.

Despite stagnation in Spearman, the hypervolume indicator (HVI) shows an increasing trend.
For CCC, HVI grows continuously; the optimizer discovers diverse specialist prompts that expand the Pareto front, even when no single prompt dominates the initialization on all four tasks.
\citet{menchaca-resendiz-klinger-2025-mopo} report a similar pattern: multi-objective prompt optimization expands the Pareto front at modest per-objective cost. In our setting, however, the per-task degradation is substantially larger when coupling is high. 

We rerun with DeepSeek v4 under MAE validation in Appendix~\ref{sec:appendix-deepseek}, where we observe better absolute improvements in Spearman through optimization due to  stronger models, but a similar trend: the \textit{improvement} through prompt optimization is most effective for single-objective optimizations.
\section{Analysis}
\label{sec:analysis}

\vspace{-0.25em}
The results in \S\ref{sec:results} establish \emph{that} multi-objective textual gradient prompt optimization lags compared to single-task; this section investigates \emph{why}.
We identify two separable failure modes: optimization-time \emph{gradient dilution} (\S\ref{sec:gradient-specificity}) and inference-time \emph{instruction interference} (\S\ref{sec:instruction-interference}). We report the evaluation prompts in Appendix~\ref{sec:diagnostic-prompts} for reproducibility.

\begin{figure}[h!]
    \centering
    \includegraphics[width=0.95\columnwidth]{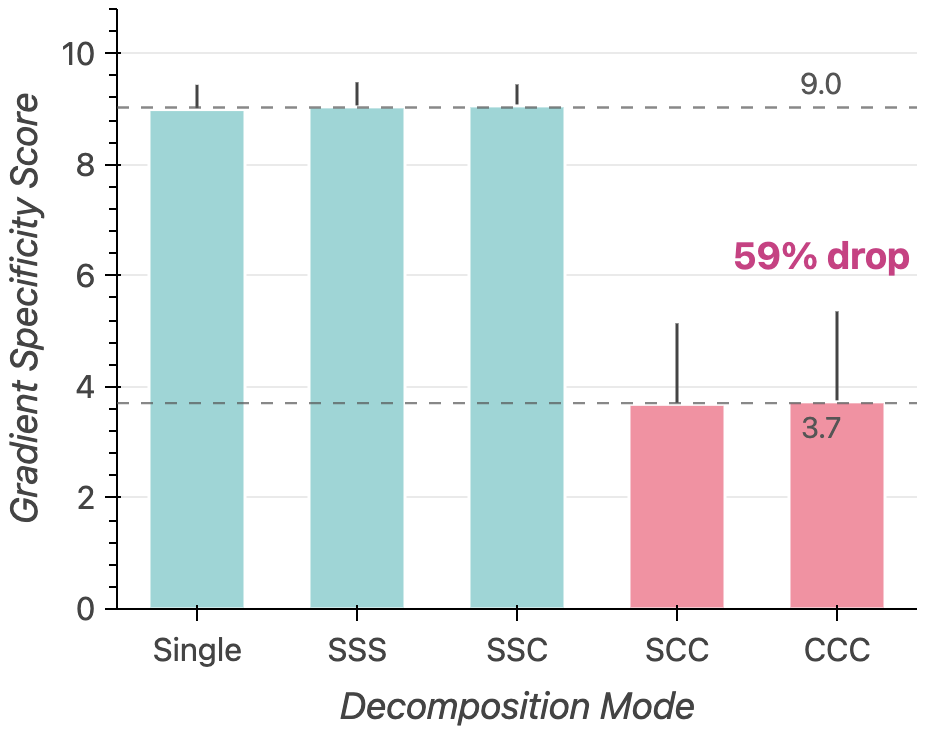}
    \caption{Gradient specificity (1 to 10 scale, higher is more task-focused) by decomposition mode.}
    \label{fig:specificity}
    \vspace{-1em}
\end{figure}

\subsection{Gradient Specificity and Dilution}
\label{sec:gradient-specificity}

To measure gradient quality directly, we evaluate each textual gradient for \emph{task-focus}: the degree to which its improvement suggestions target a single evaluation criterion rather than offering generic advice.
An LLM evaluator (Claude Sonnet~4.6, \citet{anthropic_claude_sonnet_4_6_2026}) rates each gradient on a 1 to 10 scale. A score of 10 means the gradient addresses exactly one task's rubric; a score of 1 means it's so generic it could apply to any task (see Appendix~\ref{sec:diagnostic-prompts} for the evaluation prompt). This diagnostic tests the cross-objective analogue of \citeauthor{chu2026caro}'s rule-dilution hypothesis. Concretely, we ask: does combining multiple evaluation criteria in a single gradient call dilute the criteria-specific signal?

We evaluate all gradients from steps 1 to 12 across all four modes. A sharp transition is present (Figure~\ref{fig:specificity}).
Per-task modes (Single, SSS, SSC), where each gradient LLM call processes exactly one task, achieve a mean specificity of 9.0~($\pm$0.3; Table~\ref{tab:specificity-pertask}, top rows).
Cross-task modes (SCC, CCC), where the gradient LLM processes all four tasks in a single call, drop to 3.7~($\pm$0.5; Table~\ref{tab:specificity-pertask}, bottom rows), a 59\% reduction with zero overlap between the two groups.

Our gradient dilution results extend \citeauthor{chu2026caro}'s within-criterion finding to the cross-criterion setting. \citet{chu2026caro} show that aggregating heterogeneous error modes in a single optimization step degrades rubric accuracy; we observe an analogous effect when multiple task gradients are combined in a single gradient call, degrading the per-task optimization signal.
A per-task breakdown (Table~\ref{tab:specificity-pertask}) reveals that consistency is most diluted, while coherence retains moderate focus. This possibly occurs because coherence rubrics share surface-level vocabulary with the generic ``writing quality'' feedback the gradient LLM defaults to under multi-task load.

To confirm this cliff is structural when moving from per-task to cross-task gradients, in Appendix~\ref{sec:specificity-model-swap} we switch \textit{only} the gradient LLM to a stronger LLM (\texttt{DeepSeek-V4-Pro}) while continuing to use Qwen3 for the other steps. We observe the same SSC to SCC specificity drop. Appendix~\ref{sec:deepseek-specificity} replicates the overall gradient-specificity analysis using DeepSeek for all modes, showing the same per-task vs.\ cross-task gradient dilution pattern.

\begin{table}[t!]
\centering
\small
\begin{tabular}{lcc}
\toprule
\textbf{Mode} & \textbf{MAE validation} & \textbf{No validation} \\
\midrule
Single-Task & 8.70 $\pm$ 0.47 & 8.53 $\pm$ 0.60 \\
SSS    & 8.84 $\pm$ 0.44 & 8.72 $\pm$ 0.54 \\
SSC    & 7.94 $\pm$ 0.80 & 7.83 $\pm$ 0.78 \\
SCC    & 8.08 $\pm$ 1.03 & 7.76 $\pm$ 1.49 \\
CCC    & 7.90 $\pm$ 1.62 & 8.01 $\pm$ 1.28 \\
\bottomrule
\end{tabular}
\caption{Feedback adherence scores (1 to 10 scale, mean $\pm$ std over all objectives, runs, and steps). All modes achieve high adherence, ruling out optimizer non-compliance as an explanation for multi-task optimization failure.}
\vspace{-1em}
\label{tab:adherence}
\end{table}

\subsection{Feedback Adherence}
\label{sec:feedback-adherence}

We also measure \emph{feedback adherence}: the degree to which the optimizer LLM incorporates the textual gradient into its instruction edits (Table~\ref{tab:adherence}). We prompt Claude Sonnet~4.6 evaluator to output an adherence score on a 10-point scale (refer Appendix~\ref{sec:diagnostic-prompts} for the prompt).

Across all modes and validation settings, adherence is uniformly high (7.8 to 8.8 on a 10-point scale), confirming that the optimizer faithfully implements whatever gradient it receives, even when those suggestions are generic rather than criterion-specific. This indicates the ceiling to multi-objective judge prompt optimization is gradient quality rather than optimizer compliance.

\begin{table}[t!]
\centering
\small
\begin{tabular}{l cccc c}
\toprule
\textbf{Method} & \textbf{Fl.} & \textbf{Rel.} & \textbf{Coh.} & \textbf{Con.} & \textbf{Avg $\rho$} \\
\midrule
Initial          & \textbf{0.366} & 0.208 & 0.308 & 0.256 & 0.284 \\
Single-Task          & 0.350 & 0.168 & \textbf{0.338} & \textbf{0.363} & 0\textbf{.305} \\
Cherry-pick       & 0.303 & \textbf{0.257} & 0.215 & 0.105 & 0.220 \\
% Cherry-pick (OB1)          & 0.322 & 0.186 & 0.225 & 0.195 & 0.232 \\
\bottomrule
\end{tabular}
\caption{Oracle cherry-pick experiment (MAE validation). For each task, the single-task instruction with the highest test-set metric is selected and combined into one multi-task prompt. Despite oracle selection, combined instructions degrade below the individually optimized single-task performance, demonstrating inference-time instruction interference. We report both task-level and average $\rho$ for the multi-task prompt.}
\label{tab:cherrypick}
\vspace{-1em}
\end{table}

\subsection{Inference-Time Instruction Interference}
\label{sec:instruction-interference}

Gradient dilution explains why SCC and CCC fail to optimize effectively, but it does not explain why SSS and SSC optimizations also stagnate.
These per-task modes produce highly specific gradients and high-adherence prompt updates, yet still fail to improve over the initialization.
To diagnose this, we design an oracle cherry-pick experiment that isolates the \emph{inference-time} component of multi-task failure.
For each task, we select the single best instruction across all single-task optimization runs (the instruction with the highest held-out Spearman~$\rho$ for that task), then combine these four oracle-optimal instructions into one multi-task prompt and evaluate on the full test set.

We see that oracle-optimal instructions \emph{degrade} when combined (Table~\ref{tab:cherrypick}), achieving 0.22 average Spearman, a drop of 0.085 from the single-task optimized performance of 0.305. In Appendix~\ref{sec:appendix-cherrypick-all} we explore more variants by using different metrics to select the best instruction per task, and observe the same effect. These instructions each individually outperform the baseline on their respective tasks, but combining them produces performance strictly worse than the generic initial prompt.

The primary mechanism appears to be instruction-length asymmetry.
Optimization produces over-specified rubrics for some tasks (the fluency instruction expands to $\sim$800 tokens with detailed scoring anchors) while leaving others under-specified (the relevance instruction remains at $\sim$4 tokens of the initial prompt).
When packed into a single prompt, verbose instructions receive disproportionate attention relative to brief ones at inference time.

This finding strengthens a result from \citet{shen2026rrd}, who observe that naive rubric construction degrades GPT-4o preference-judgment accuracy by 13 points on JudgeBench.
The degradation we observe is larger and occurs with oracle-selected rather than naively constructed instructions.
\citeauthor{shen2026rrd}'s result shows that \emph{bad} rubrics hurt. Ours shows that individually \emph{good} rubrics can hurt when combined. This implies that instruction interference is not resolvable by improving per-task optimization alone.

\section{Conclusion}
\label{sec:conclusion}

\vspace{-0.25em}
Multi-criteria textual gradient optimization for LLM judges exhibits two failure points that our decomposition study and process-level diagnostics expose.
These are systematic failures, corresponding to distinct pipeline stages and affect different decomposition modes:
(i) Gradient dilution operates at optimization time. When the gradient LLM must reconcile feedback from multiple criteria in a single call, its suggestions lose task-specificity (59\% drop; Appendix Table~\ref{tab:specificity-pertask}). The optimizer propagates the low-specificity signal through to the final prompt;
(ii) Instruction interference operates at inference time. Independently optimized per-task instructions degrade when combined into one prompt (Table~\ref{tab:cherrypick}) because instruction-length asymmetry causes verbose rubrics to receive disproportionate attention relative to brief ones.

For practitioners customizing judges to domain-specific criteria \citep{kim2024prometheus2}, these results indicate architectural changes are required before the multi-objective setting can work reliably.
Addressing either failure mode alone is insufficient.
Separate judge calls per criterion eliminates interference but multiplies inference cost.
Conflict-aware gradient resolution adapted from numerical multi-task learning~\citep{yu2020pcgrad, liu2021cagrad} could address dilution if textual gradients can be meaningfully embedded and projected.
Our proposed diagnostics (gradient specificity and feedback adherence) provide the measurement tools to evaluate either approach.

\section{Future Work}
\label{sec:future-work}

\vspace{-0.25em}
Our findings open several concrete directions to broader research on customized LLM evaluation.

\insection{Statistically reliable LLM diagnostics with PPI}
Our diagnostics are LLM-judged, introducing an evaluator-bias. Prediction-Powered Inference~\citep{angelopoulos2023ppi, boyeau2025autoeval} combines a small human-judged set with a large LLM-judged set to produce provably unbiased estimates; the hierarchical PPI extension of \citet{divekar-majumder-2026-precise} is directly applicable here, since annotations are per-gradient but the quantity of interest is the per-mode mean. This would allow us to report the specificity gap as a confidence interval and scale diagnostics to hundreds of runs without scaling human annotation linearly.

\insection{Synthetic task generation for aligned criteria}
A new direction is to \emph{synthesize the criteria themselves} rather than treat them as fixed. Persona-driven synthesis at billion-person scale~\citep{chan2024scaling, yu2023large} and instruction-data generation with controlled diversity~\citep{divekar-durrett-2024-synthesizrr, kowshik-etal-2024-corrsynth} can produce synthetic objectives that are complementary for optimization. If the synthesized tasks are mutually aligned by construction, the combined gradient should suffer less semantic drift, raising gradient focus and potentially mitigating both failure modes at their source.

\insection{Multi-objective critics for agentic workflows}
Multi-objective judge prompts could serve as critics in agentic LLM systems, where a critic must track several quality dimensions of tool-use trajectories at once~\citep{ding2026adarubric, chuang2026scalableverifiereward, rudman2026vesta}. Applying our optimization pipeline to such critics raises an open question: do gradient dilution and instruction interference still emerge when criteria are tool-grounded and partially verifiable, or does verifiability yield more robust gradients? Our diagnostics provide a principled way to measure this.

\insection{Mitigations}
Our diagnostics motivate concrete mitigations. For \emph{gradient dilution}, a specificity-aware router could fallback to per-task gradient LLM when a multi-task LLM specificity drops below threshold, capturing CCC's hypervolume without losing task-focus. For \emph{instruction interference}, we propose two avenues: (i) next-token attention masking for per-criterion output generation (eliminating interference as no cost), and (ii) length-aware instruction synthesis that normalizes rubric length during optimization so no single criterion dominates the attention budget.

\section*{Limitations}
\label{sec:limitations}

\vspace{-0.25em}
Our experiments are scoped to \SummEval{} as it provides expert human annotations on four clearly separable criteria. Other benchmarks would validate our identified failure modes: \BRIGHTER{}~\citep{muhammad-etal-2025-brighter} tests whether dilution scales with task count and crosses language boundaries; ASAP++~\citep{mathias-bhattacharyya-2018-asap} (per-trait essay-grading rubrics) tests whether the SCC to CCC gradient specificity cliff transfers beyond summarization; EMSCAD~\citep{vidros2017emscad}, GitBugs~\citep{patil2025gitbugs}, test heterogeneous classification criteria rather than ordinal quality scales.

Other prompt optimization paradigms may exhibit different multi-task dynamics, though our diagnostics are algorithm-agnostic and can be applied to any textual gradient approach. 

Our sample size ($N\!=\!3$ runs) limits statistical power, and we restrict our claims to effect sizes that are robust at this sample size (e.g., the 59\% specificity drop and $-0.085$ cherry-pick degradation). Finally, gradient specificity and feedback adherence are scored by an LLM evaluator, which introduces a potential confound.

\section*{Ethics Statement}
\label{sec:ethics}

Optimized judge prompts inherit biases present in the underlying LLMs and in the human annotations used for evaluation; practitioners should audit optimized prompts before deploying them in sensitive evaluation contexts (e.g., content moderation or hiring).
Our code and diagnostics will be released under an open-source license to support reproducibility.

% Bibliography
\bibliography{custom}

@inproceedings{yang2024opro,
  author = {Yang, Chengrun and Wang, Xuezhi and Lu, Yifeng and Liu, Hanxiao and Le, Quoc V and Zhou, Denny and Chen, Xinyun},
  booktitle = {International Conference on Learning Representations},
  editor = {B. Kim and Y. Yue and S. Chaudhuri and K. Fragkiadaki and M. Khan and Y. Sun},
  pages = {12028--12068},
  title = {Large Language Models as Optimizers},
  url = {https://proceedings.iclr.cc/paper_files/paper/2024/file/3339f19c5fcee3ad74502947a32be9e6-Paper-Conference.pdf},
  volume = {2024},
  year = {2024}
}

@article{yuksekgonul2025textgrad,
  title={Optimizing generative {AI} by backpropagating language model feedback},
  author={Y\"{u}ksekg\"{o}n\"{u}l, Mert and Bianchi, Federico and Boen, Joseph and Liu, Sheng and Lu, Pan and Huang, Zhi and Guestrin, Carlos and Zou, James},
  journal={Nature},
  volume={639},
  number={8055},
  pages={609--616},
  year={2025},
  doi={10.1038/s41586-025-08661-4},
  url={https://www.nature.com/articles/s41586-025-08661-4}
}

@inproceedings{tang2025gpo,
  author = {Tang, Xinyu and Wang, Xiaolei and Zhao, Wayne Xin and Lu, Siyuan and Li, Yaliang and Wen, Ji-Rong},
  title = {Unleashing the Potential of Large Language Models as Prompt Optimizers: Analogical Analysis with Gradient-based Model Optimizers},
  year = {2025},
  url = {https://doi.org/10.1609/aaai.v39i24.34713},
  doi = {10.1609/aaai.v39i24.34713},
  booktitle = {Proceedings of the Thirty-Ninth AAAI Conference on Artificial Intelligence}
}

@inproceedings{pryzant-etal-2023-automatic,
  title = "Automatic Prompt Optimization with ``Gradient Descent'' and Beam Search",
  author = "Pryzant, Reid and
    Iter, Dan and
    Li, Jerry and
    Lee, Yin and
    Zhu, Chenguang and
    Zeng, Michael",
  editor = "Bouamor, Houda and
    Pino, Juan and
    Bali, Kalika",
  booktitle = "Proceedings of the 2023 Conference on Empirical Methods in Natural Language Processing",
  month = dec,
  year = "2023",
  address = "Singapore",
  publisher = "Association for Computational Linguistics",
  url = "https://aclanthology.org/2023.emnlp-main.494/",
  doi = "10.18653/v1/2023.emnlp-main.494",
  pages = "7957--7968"
}

@inproceedings{zhou2023ape,
  title={Large Language Models are Human-Level Prompt Engineers},
  author={Yongchao Zhou and Andrei Ioan Muresanu and Ziwen Han and Keiran Paster and Silviu Pitis and Harris Chan and Jimmy Ba},
  booktitle={The Eleventh International Conference on Learning Representations},
  year={2023},
  url={https://openreview.net/forum?id=92gvk82DE-}
}

@inproceedings{menchaca-resendiz-klinger-2025-mopo,
  title = "{MOPO}: Multi-Objective Prompt Optimization for Affective Text Generation",
  author = "Menchaca Resendiz, Yarik and
    Klinger, Roman",
  editor = "Rambow, Owen and
    Wanner, Leo and
    Apidianaki, Marianna and
    Al-Khalifa, Hend and
    Eugenio, Barbara Di and
    Schockaert, Steven",
  booktitle = "Proceedings of the 31st International Conference on Computational Linguistics",
  month = jan,
  year = "2025",
  address = "Abu Dhabi, UAE",
  publisher = "Association for Computational Linguistics",
  url = "https://aclanthology.org/2025.coling-main.375/",
  pages = "5588--5606"
}

@inproceedings{jafari-etal-2024-morl,
  title = "{MORL}-Prompt: An Empirical Analysis of Multi-Objective Reinforcement Learning for Discrete Prompt Optimization",
  author = "Jafari, Yasaman and
    Mekala, Dheeraj and
    Yu, Rose and
    Berg-Kirkpatrick, Taylor",
  editor = "Al-Onaizan, Yaser and
    Bansal, Mohit and
    Chen, Yun-Nung",
  booktitle = "Findings of the Association for Computational Linguistics: EMNLP 2024",
  month = nov,
  year = "2024",
  address = "Miami, Florida, USA",
  publisher = "Association for Computational Linguistics",
  url = "https://aclanthology.org/2024.findings-emnlp.577/",
  doi = "10.18653/v1/2024.findings-emnlp.577",
  pages = "9878--9889"
}

@inproceedings{zhao2025paretoprompt,
  title={Pareto Prompt Optimization},
  author={Zhao, Guang and Yoon, Byung-Jun and Park, Gilchan and Jha, Shantenu and Yoo, Shinjae and Qian, Xiaoning},
  booktitle={The Thirteenth International Conference on Learning Representations},
  year={2025},
  url={https://openreview.net/forum?id=HGCk5aaSvE}
}

@inproceedings{guo2024connecting,
  title={Connecting Large Language Models with Evolutionary Algorithms Yields Powerful Prompt Optimizers},
  author={Qingyan Guo and Rui Wang and Junliang Guo and Bei Li and Kaitao Song and Xu Tan and Guoqing Liu and Jiang Bian and Yujiu Yang},
  booktitle={The Twelfth International Conference on Learning Representations},
  year={2024},
  url={https://openreview.net/forum?id=ZG3RaNIsO8}
}

@inproceedings{baumann2024emoprompts,
  title={Evolutionary Multi-Objective Optimization of Large Language Model Prompts for Balancing Sentiments},
  author={Baumann, Jill and Kramer, Oliver},
  booktitle={Applications of Evolutionary Computation (EvoApplications)},
  pages={212--224},
  year={2024},
  publisher={Springer},
  doi={10.1007/978-3-031-56855-8_13}
}

@inproceedings{agrawal2026gepa,
  title={{GEPA}: Reflective Prompt Evolution Can Outperform Reinforcement Learning},
  author={Agrawal, Lakshya A. and Tan, Shangyin and Soylu, Dilara and Ziems, Noah and Khare, Rishi and Opsahl-Ong, Krista and Singhvi, Arnav and Shandilya, Herumb and Ryan, Michael J. and Jiang, Meng and Potts, Christopher and Sen, Koushik and Dimakis, Alexandros G. and Stoica, Ion and Klein, Daniel and Zaharia, Matei and Khattab, Omar},
  booktitle={The Fourteenth International Conference on Learning Representations},
  year={2026},
  url={https://openreview.net/forum?id=RQm2KQTM5r}
}

@inproceedings{sener2018moo,
  author = {Sener, Ozan and Koltun, Vladlen},
  booktitle = {Advances in Neural Information Processing Systems},
  editor = {S. Bengio and H. Wallach and H. Larochelle and K. Grauman and N. Cesa-Bianchi and R. Garnett},
  pages = {},
  publisher = {Curran Associates, Inc.},
  title = {Multi-Task Learning as Multi-Objective Optimization},
  url = {https://proceedings.neurips.cc/paper_files/paper/2018/file/432aca3a1e345e339f35a30c8f65edce-Paper.pdf},
  volume = {31},
  year = {2018}
}

@article{deb2002nsga2,
  title={A fast and elitist multiobjective genetic algorithm: {NSGA-II}},
  author={Deb, Kalyanmoy and Pratap, Amrit and Agarwal, Sameer and Meyarivan, T.},
  journal={IEEE transactions on evolutionary computation},
  volume={6},
  number={2},
  pages={182--197},
  year={2002}
}

@inproceedings{zheng2023judging,
  title={Judging {LLM}-as-a-Judge with {MT}-Bench and Chatbot Arena},
  author={Lianmin Zheng and Wei-Lin Chiang and Ying Sheng and Siyuan Zhuang and Zhanghao Wu and Yonghao Zhuang and Zi Lin and Zhuohan Li and Dacheng Li and Eric Xing and Hao Zhang and Joseph E. Gonzalez and Ion Stoica},
  booktitle={Thirty-seventh Conference on Neural Information Processing Systems Datasets and Benchmarks Track},
  year={2023},
  url={https://openreview.net/forum?id=uccHPGDlao}
}

@article{fabbri-etal-2021-summeval,
  title = "{S}umm{E}val: Re-evaluating Summarization Evaluation",
  author = "Fabbri, Alexander R. and
    Kry{\'s}ci{\'n}ski, Wojciech and
    McCann, Bryan and
    Xiong, Caiming and
    Socher, Richard and
    Radev, Dragomir",
  editor = "Roark, Brian and
    Nenkova, Ani",
  journal = "Transactions of the Association for Computational Linguistics",
  volume = "9",
  year = "2021",
  address = "Cambridge, MA",
  publisher = "MIT Press",
  url = "https://aclanthology.org/2021.tacl-1.24/",
  doi = "10.1162/tacl_a_00373",
  pages = "391--409"
}

@inproceedings{yu2020pcgrad,
  author = {Yu, Tianhe and Kumar, Saurabh and Gupta, Abhishek and Levine, Sergey and Hausman, Karol and Finn, Chelsea},
  booktitle = {Advances in Neural Information Processing Systems},
  editor = {H. Larochelle and M. Ranzato and R. Hadsell and M.F. Balcan and H. Lin},
  pages = {5824--5836},
  publisher = {Curran Associates, Inc.},
  title = {Gradient Surgery for Multi-Task Learning},
  url = {https://proceedings.neurips.cc/paper_files/paper/2020/file/3fe78a8acf5fda99de95303940a2420c-Paper.pdf},
  volume = {33},
  year = {2020}
}

@inproceedings{liu2021cagrad,
  title={Conflict-Averse Gradient Descent for Multi-task Learning},
  author={Liu, Bo and Liu, Xingchao and Jin, Xiaojie and Stone, Peter and Liu, Qiang},
  booktitle={Advances in Neural Information Processing Systems},
  pages={18878--18890},
  year={2021}
}

@inproceedings{muhammad-etal-2025-brighter,
    title = "{BRIGHTER}: {BRI}dging the Gap in Human-Annotated Textual Emotion Recognition Datasets for 28 Languages",
    author = "Muhammad, Shamsuddeen Hassan  and
      Ousidhoum, Nedjma  and
      Abdulmumin, Idris  and
      Wahle, Jan Philip  and
      Ruas, Terry  and
      Beloucif, Meriem  and
      de Kock, Christine  and
      Surange, Nirmal  and
      Teodorescu, Daniela  and
      Ahmad, Ibrahim Said  and
      Adelani, David Ifeoluwa  and
      Aji, Alham Fikri  and
      Ali, Felermino D. M. A.  and
      Alimova, Ilseyar  and
      Araujo, Vladimir  and
      Babakov, Nikolay  and
      Baes, Naomi  and
      Bucur, Ana-Maria  and
      Bukula, Andiswa  and
      Cao, Guanqun  and
      Tufi{\~n}o, Rodrigo  and
      Chevi, Rendi  and
      Chukwuneke, Chiamaka Ijeoma  and
      Ciobotaru, Alexandra  and
      Dementieva, Daryna  and
      Gadanya, Murja Sani  and
      Geislinger, Robert  and
      Gipp, Bela  and
      Hourrane, Oumaima  and
      Ignat, Oana  and
      Lawan, Falalu Ibrahim  and
      Mabuya, Rooweither  and
      Mahendra, Rahmad  and
      Marivate, Vukosi  and
      Panchenko, Alexander  and
      Piper, Andrew  and
      Ferreira, Charles Henrique Porto  and
      Protasov, Vitaly  and
      Rutunda, Samuel  and
      Shrivastava, Manish  and
      Udrea, Aura Cristina  and
      Wanzare, Lilian Diana Awuor  and
      Wu, Sophie  and
      Wunderlich, Florian Valentin  and
      Zhafran, Hanif Muhammad  and
      Zhang, Tianhui  and
      Zhou, Yi  and
      Mohammad, Saif M.",
    editor = "Che, Wanxiang  and
      Nabende, Joyce  and
      Shutova, Ekaterina  and
      Pilehvar, Mohammad Taher",
    booktitle = "Proceedings of the 63rd Annual Meeting of the Association for Computational Linguistics (Volume 1: Long Papers)",
    month = jul,
    year = "2025",
    address = "Vienna, Austria",
    publisher = "Association for Computational Linguistics",
    url = "https://aclanthology.org/2025.acl-long.436/",
    doi = "10.18653/v1/2025.acl-long.436",
    pages = "8895--8916",
    ISBN = "979-8-89176-251-0",
}

@inproceedings{mathias-bhattacharyya-2018-asap,
  author       = {Sandeep Mathias and
                  Pushpak Bhattacharyya},
  title        = {{ASAP++:} Enriching the {ASAP} Automated Essay Grading Dataset with
                  Essay Attribute Scores},
  booktitle    = {Proceedings of the Eleventh International Conference on Language Resources
                  and Evaluation, {LREC} 2018, Miyazaki, Japan, May 7-12, 2018},
  publisher    = {European Language Resources Association {(ELRA)}},
  year         = {2018},
  url          = {http://www.lrec-conf.org/proceedings/lrec2018/summaries/373.html}
}

@article{vidros2017emscad,
  author       = {Sokratis Vidros and
                  Constantinos Kolias and
                  Georgios Kambourakis and
                  Leman Akoglu},
  title        = {Automatic Detection of Online Recruitment Frauds: Characteristics,
                  Methods, and a Public Dataset},
  journal      = {Future Internet},
  volume       = {9},
  number       = {1},
  pages        = {6},
  year         = {2017},
  doi          = {10.3390/fi9010006},
  url          = {https://doi.org/10.3390/fi9010006}
}

@misc{patil2025gitbugs,
      title={GitBugs: Bug Reports for Duplicate Detection, Retrieval Augmented Generation, Triage, and More}, 
      author={Avinash Patil and Siru Tao and Aryan Jadon},
      year={2026},
      eprint={2504.09651},
      archivePrefix={arXiv},
      primaryClass={cs.SE},
      url={https://arxiv.org/abs/2504.09651}, 
}

@inproceedings{divekar-durrett-2024-synthesizrr,
  author       = {Abhishek Divekar and
                  Greg Durrett},
  title        = {SynthesizRR: Generating Diverse Datasets with Retrieval Augmentation},
  booktitle    = {Proceedings of the 2024 Conference on Empirical Methods in Natural
                  Language Processing, {EMNLP} 2024, Miami, FL, USA, November 12-16,
                  2024},
  pages        = {19200--19227},
  publisher    = {Association for Computational Linguistics},
  year         = {2024},
  url          = {https://aclanthology.org/2024.emnlp-main.1071/},
  doi          = {10.18653/v1/2024.emnlp-main.1071}
}

@inproceedings{kowshik-etal-2024-corrsynth,
  author       = {Suhas S. Kowshik and
                  Abhishek Divekar and
                  Vijit Malik},
  title        = {{CorrSynth}: {A} Correlated Sampling Method for Diverse Dataset Generation
                  from {LLMs}},
  booktitle    = {Proceedings of the 2024 Conference on Empirical Methods in Natural
                  Language Processing, {EMNLP} 2024, Miami, FL, USA, November 12-16,
                  2024},
  pages        = {16076--16095},
  publisher    = {Association for Computational Linguistics},
  year         = {2024},
  url          = {https://aclanthology.org/2024.emnlp-main.899/},
  doi          = {10.18653/v1/2024.emnlp-main.899}
}

@article{angelopoulos2023ppi,
  author       = {Angelopoulos, Anastasios N. and Bates, Stephen and Fannjiang, Clara and Jordan, Michael I. and Zrnic, Tijana},
  title        = {Prediction-Powered Inference},
  journal      = {Science},
  volume       = {382},
  number       = {6671},
  pages        = {669--674},
  year         = {2023},
  doi          = {10.1126/science.adi6000},
  url          = {https://doi.org/10.1126/science.adi6000}
}

@inproceedings{divekar-majumder-2026-precise,
  author       = {Abhishek Divekar and
                  Anirban Majumder},
  title        = {{PRECISE}: Reducing the Bias of {LLM} Evaluations Using Prediction-Powered Ranking Estimation},
  booktitle    = {Proceedings of the AAAI Conference on Artificial Intelligence (IAAI Track)},
  volume       = {40},
  pages        = {39929--39938},
  year         = {2026},
  doi          = {10.1609/aaai.v40i47.41427},
  url          = {https://doi.org/10.1609/aaai.v40i47.41427}
}

@article{melcer2025textgrad,
  title={Textual Gradients are a Flawed Metaphor for Automatic Prompt Optimization},
  author={Melcer, Daniel and Chen, Qi and Chiang, Wen-Hao and Garg, Shweta and Garg, Pranav and Bock, Christian},
  journal={arXiv preprint arXiv:2512.13598},
  year={2025}
}

@inproceedings{kim2024prometheus2,
  title={Prometheus 2: An Open Source Language Model Specialized in Evaluating Other Language Models},
  author={Kim, Seungone and Suk, Juyoung and Longpre, Shayne and Lin, Bill Yuchen and Shin, Jamin and Welleck, Sean and Neubig, Graham and Lee, Moontae and Lee, Kyungjae and Seo, Minjoon},
  booktitle={Proceedings of the 2024 Conference on Empirical Methods in Natural Language Processing},
  pages={4334--4353},
  year={2024},
  doi={10.18653/v1/2024.emnlp-main.248},
  url={https://aclanthology.org/2024.emnlp-main.248/}
}

@inproceedings{whitehouse2026j1,
  title={{J1}: Incentivizing Thinking in {LLM}-as-a-Judge via Reinforcement Learning},
  author={Whitehouse, Chenxi and Wang, Tianlu and Yu, Ping and Li, Xian and Weston, Jason and Kulikov, Ilia and Saha, Swarnadeep},
  booktitle={The Fourteenth International Conference on Learning Representations},
  year={2026},
  url={https://openreview.net/forum?id=dnJEHl6DI1}
}

@article{yang2026fairjudge,
  title={{FairJudge}: An Adaptive, Debiased, and Consistent {LLM}-as-a-Judge},
  author={Yang, Bo and Feng, Lanfei and Chen, Yunkui and Zhang, Yu and Xu, Xiao and Li, Shijian},
  journal={arXiv preprint arXiv:2602.06625},
  year={2026}
}

@inproceedings{wang2025djpo,
  title={Direct Judgement Preference Optimization},
  author={Wang, Peifeng and Xu, Austin and Zhou, Yilun and Xiong, Caiming and Joty, Shafiq},
  booktitle={Proceedings of the 2025 Conference on Empirical Methods in Natural Language Processing},
  pages={1979--2009},
  year={2025},
  doi={10.18653/v1/2025.emnlp-main.103},
  url={https://aclanthology.org/2025.emnlp-main.103/}
}

@article{chu2026caro,
  title={Confusion-Aware Rubric Optimization for {LLM}-based Automated Grading},
  author={Chu, Yucheng and Li, Hang and Yang, Kaiqi and Copur-Gencturk, Yasemin and Krajcik, Joseph and Shin, Namsoo and Tang, Jiliang},
  journal={arXiv preprint arXiv:2603.00451},
  year={2026}
}

@article{shen2026rrd,
  title={Rethinking Rubric Generation for Improving {LLM} Judge and Reward Modeling for Open-ended Tasks},
  author={Shen, William F. and Qiu, Xinchi and Whitehouse, Chenxi and Alazraki, Lisa and Goel, Shashwat and Barbieri, Francesco and Willi, Timon and Mathur, Akhil and Leontiadis, Ilias},
  journal={arXiv preprint arXiv:2602.05125},
  year={2026}
}

@article{sharma2026mpo,
  title={Modular Prompt Optimization: Optimizing Structured Prompts with Section-Local Textual Gradients},
  author={Sharma, Prith and Henley, Austin Z.},
  journal={arXiv preprint arXiv:2601.04055},
  year={2026}
}

@inproceedings{han2025mapgd,
  title={{MAPGD}: Multi-Agent Prompt Gradient Descent for Collaborative Prompt Optimization},
  author={Han, Yichen and Liu, Bojun and Zhou, Zhengpeng and Liu, Guanyu and Zhang, Zeng and Yang, Yang and Wang, Wenli and Shi, Isaac and Yunyan and He, Lewei and Shi, Tianyu},
  booktitle={NeurIPS 2025 Workshop on Scaling Environments for Agents},
  year={2025},
  url={https://openreview.net/forum?id=FywYwwH5z9}
}

@misc{deepseekai2026deepseekv4,
      title={DeepSeek-V4: Towards Highly Efficient Million-Token Context Intelligence},
      author={DeepSeek-AI},
      year={2026},
}

@misc{anthropic_claude_sonnet_4_6_2026,
  author       = {{Anthropic}},
  title        = {Introducing Claude Sonnet 4.6},
  howpublished = {Anthropic Blog},
  year         = {2026},
  month        = {February},
  url          = {https://www.anthropic.com/news/claude-sonnet-4-6}
}

@article{Liu2023GEvalNE,
  title={G-Eval: NLG Evaluation using GPT-4 with Better Human Alignment},
  author={Yang Liu and Dan Iter and Yichong Xu and Shuo Wang and Ruochen Xu and Chenguang Zhu},
  journal={ArXiv},
  year={2023},
  volume={abs/2303.16634},
  url={https://api.semanticscholar.org/CorpusID:257804696}
}

@misc{yang2025qwen3technicalreport,
      title={Qwen3 Technical Report}, 
      author={An Yang and Anfeng Li and Baosong Yang and Beichen Zhang and Binyuan Hui and Bo Zheng and Bowen Yu and Chang Gao and Chengen Huang and Chenxu Lv and Chujie Zheng and Dayiheng Liu and Fan Zhou and Fei Huang and Feng Hu and Hao Ge and Haoran Wei and Huan Lin and Jialong Tang and Jian Yang and Jianhong Tu and Jianwei Zhang and Jianxin Yang and Jiaxi Yang and Jing Zhou and Jingren Zhou and Junyang Lin and Kai Dang and Keqin Bao and Kexin Yang and Le Yu and Lianghao Deng and Mei Li and Mingfeng Xue and Mingze Li and Pei Zhang and Peng Wang and Qin Zhu and Rui Men and Ruize Gao and Shixuan Liu and Shuang Luo and Tianhao Li and Tianyi Tang and Wenbiao Yin and Xingzhang Ren and Xinyu Wang and Xinyu Zhang and Xuancheng Ren and Yang Fan and Yang Su and Yichang Zhang and Yinger Zhang and Yu Wan and Yuqiong Liu and Zekun Wang and Zeyu Cui and Zhenru Zhang and Zhipeng Zhou and Zihan Qiu},
      year={2025},
      eprint={2505.09388},
      archivePrefix={arXiv},
      primaryClass={cs.CL},
      url={https://arxiv.org/abs/2505.09388}, 
}

@article{yu2023large,
  title={Large Language Model as Attributed Training Data Generator: A Tale of Diversity and Bias},
  author={Yu, Yue and Zhuang, Yuchen and Zhang, Jieyu and Meng, Yu and Ratner, Alexander J. and Krishna, Ranjay and Shen, Jiaming and Zhang, Chao},
  journal={arXiv preprint arXiv:2306.15895},
  year={2023},
  url={https://arxiv.org/abs/2306.15895}
}

@article{chan2024scaling,
  title={Scaling Synthetic Data Creation with 1,000,000,000 Personas},
  author={Chan, Xin and Wang, Xiaoyang and Yu, Dian and Mi, Haitao and Yu, Dong},
  journal={arXiv preprint arXiv:2406.20094},
  year={2024},
  url={https://arxiv.org/abs/2406.20094}
}

@misc{rudman2026vesta,
  title={VESTA: Visual Exploration with Statistical Tool Agents},
  author={Rudman, William and Divekar, Abhishek and Jain, Kanishk and Joseph, Sebastian and Offner, Stella S. R. and Lease, Matthew and Mahowald, Kyle and Durrett, Greg and Li, Junyi Jessy},
  year={2026},
  eprint={2606.00384},
  archivePrefix={arXiv},
  primaryClass={cs.AI},
  url={https://arxiv.org/abs/2606.00384}
}

@misc{ding2026adarubric,
      title={AdaRubric: Task-Adaptive Rubrics for Reliable LLM Agent Evaluation and Reward Learning},
      author={Liang Ding},
      year={2026},
      eprint={2603.21362},
      archivePrefix={arXiv},
      primaryClass={cs.AI},
      url={https://arxiv.org/abs/2603.21362}
}

@misc{chuang2026scalableverifiereward,
      title={Toward Scalable Verifiable Reward: Proxy State-Based Evaluation for Multi-turn Tool-Calling LLM Agents},
      author={Yun-Shiuan Chuang and Chaitanya Kulkarni and Alec Chiu and Avinash Thangali and Zijie Pan and Shivani Shekhar and Yirou Ge and Yixi Li and Uma Kona and Linsey Pang and Prakhar Mehrotra},
      year={2026},
      eprint={2602.16246},
      archivePrefix={arXiv},
      primaryClass={cs.AI},
      url={https://arxiv.org/abs/2602.16246}
}

@inproceedings{boyeau2025autoeval,
  title={AutoEval Done Right: Using Synthetic Data for Model Evaluation},
  author={Boyeau, Pierre and Angelopoulos, Anastasios Nikolas and Li, Tianle and Yosef, Nir and Malik, Jitendra and Jordan, Michael I.},
  booktitle={Forty-second International Conference on Machine Learning},
  year={2025},
  url={https://openreview.net/forum?id=S8kbmk12Oo}
}

\appendix

\section{Per-Task Gradient Specificity}
\label{sec:appendix-specificity-pertask}

Table~\ref{tab:specificity-pertask} shows gradient specificity broken down by task for the combined-gradient modes (SCC, CCC).
Averaged across SCC and CCC, consistency is the most diluted dimension (specificity 2.5), while coherence retains moderate focus (5.0).
This suggests that the gradient LLM's attention is not uniformly distributed across tasks when processing them simultaneously.

\begin{table}[h]
\centering
\small
\begin{tabular}{lccccc}
\toprule
\textbf{Mode} & \textbf{Fl.} & \textbf{Rel.} & \textbf{Coh.} & \textbf{Con.} & \textbf{Avg} \\
\midrule
\multicolumn{6}{c}{\textsc{Per-task gradient modes}} \\
[0.5ex]
Single & 8.9 & 8.9 & 9.1 & 9.0 &  9.0 \\
SSS    & 9.0 & 9.0 & 9.1 & 9.0 &  9.0 \\
SSC    & 9.0 & 9.1 & 9.1 & 9.0 &  9.0 \\
[1.0ex]
\multicolumn{6}{c}{\textsc{Cross-task gradient modes}} \\
[0.5ex]
SCC    & 3.0 & 4.3 & 4.8 & 2.6 & 3.7 \\ 
CCC    & 3.2 & 4.3 & 5.1 & 2.4 & 3.8 \\ 
\bottomrule
\end{tabular}
\caption{Gradient specificity by task (mean over both val settings, all runs and steps). Per-task modes maintain uniformly high specificity. In all-task modes, consistency is most diluted and coherence retains moderate focus.}
\label{tab:specificity-pertask}
\end{table}

\section{Cherry-Pick Experiment: All Variants}
\label{sec:appendix-cherrypick-all}

Table~\ref{tab:cherrypick-all} shows all six cherry-pick variants (three selection metrics and two validation settings).
All variants degrade below the initial generic baseline, confirming that inference-time instruction interference is robust across metric choices.

\begin{table}[h]
\centering
\small
\begin{tabular}{llc}
\toprule
\textbf{Validation} & \textbf{Selection metric} & \textbf{Avg $\rho$} \\
\midrule
\multicolumn{2}{l}{Initial prompt (generic)} & 0.284 \\
\midrule
MAE  & Spearman & 0.220 \\
MAE  & Off-by-one      & 0.232 \\
MAE  & MAE      & 0.231 \\
[0.5ex]
None & Spearman & 0.120 \\
None & Off-by-one      & 0.200 \\
None & MAE      & 0.172 \\
\bottomrule
\end{tabular}
\caption{All cherry-pick variants. Every oracle combination degrades below the initial generic baseline. The worst case (val=none, Spearman selection) additionally shows zero fluency correlation.}
\label{tab:cherrypick-all}
\end{table}

\section{Gradient Specificity Under Gradient-Model Swap}
\label{sec:specificity-model-swap}

% We swap the gradient LLM to a different backbone (\texttt{DeepSeek-V4-Pro}) while keeping the task, loss, and optimizer LLMs in the Qwen family and holding the rest of the pipeline fixed (8 optimization steps). Figure~\ref{fig:specificity-swap-bars} shows SSC remains high (\(8.82 \pm 0.10\)) while SCC drops to \(4.22 \pm 0.26\) (a \textbf{52\%} reduction), indicating that gradient dilution persists under a cross-family gradient model.

\begin{figure}[h!]
    \centering
    \includegraphics[width=0.95\linewidth]{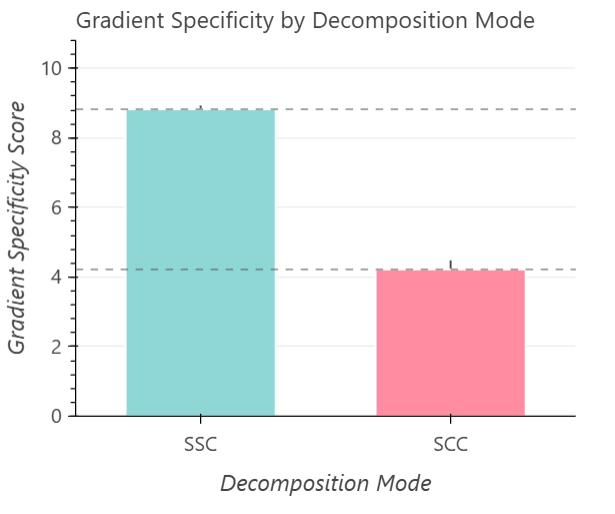}
    \caption{Gradient specificity for SSC vs.\ SCC after swapping the gradient LLM to DeepSeek-V4-Pro while keeping the other stages on Qwen3. SSC remains high ($8.82 \pm 0.10$) while SCC drops to $4.22 \pm 0.26$, a 52\% reduction.}
    \label{fig:specificity-swap-bars}
\end{figure}

We swap the gradient LLM backbone to \texttt{DeepSeek-V4-Pro} \citep{deepseekai2026deepseekv4} while keeping the task, loss, and optimizer LLMs in the Qwen3 family and holding the rest of the pipeline fixed. As shown in Figure~\ref{fig:specificity-swap-bars}, SSC remains high ($8.82 \pm 0.10$) while SCC drops sharply ($4.22 \pm 0.26$, a 52\% reduction), indicating that gradient dilution persists under a cross-family gradient model.

\begin{figure*}[t!]
    \centering
    \includegraphics[width=0.95\textwidth]{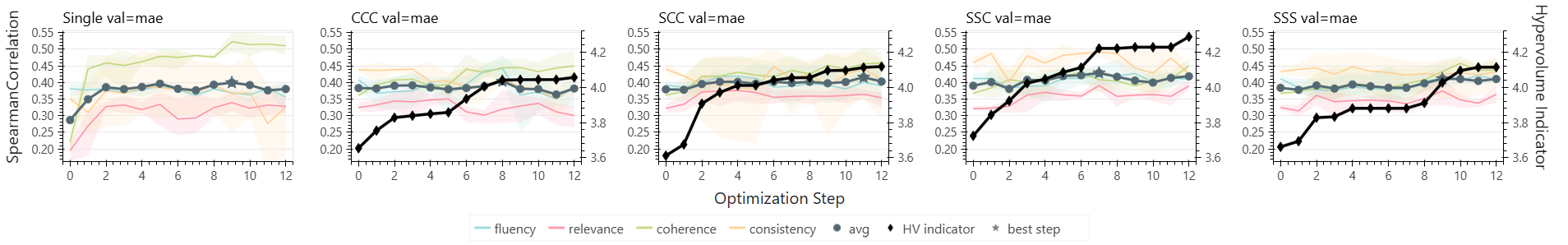}
    \caption{Per-task Spearman $\rho$ for each optimization steps on \SummEval{} with DeepSeek v4. Notation same as Figure~\ref{fig:trajectories}.}
    \label{fig:deepseek-trajectories}
\end{figure*}

\begin{table}[h!]   
\centering
\small
\setlength{\tabcolsep}{4pt}
\begin{tabular}{lcccc}
\toprule
\textbf{Mode} & \textbf{Initial} & \textbf{Best (step)} & \textbf{$\Delta$} & \textbf{HVI} \\
\midrule
\textsc{Single-Task} & 0.287 & 0.403 (9)               & \textbf{+0.117} & ---   \\
[0.5ex]
\textsc{SSS}    & 0.384 & 0.421 (9)               & +0.037 & 4.264 \\
\textsc{SSC}    & \textbf{0.390} & \textbf{0.429} (7) & +0.039 & \textbf{4.436} \\
\textsc{SCC}    & 0.379 & 0.418 (11)              & +0.039 & 4.331 \\
\textsc{CCC}    & 0.383 & 0.409 (7)               & +0.026 & 4.282 \\
\bottomrule
\end{tabular}
\caption{Main results under the MAE validation gate for the DeepSeek backbone. We measure the task-averaged Spearman correlation (mean over $N\!=\!3$ runs). Notation same as Table~\ref{tab:main-results}.}
\label{tab:deepseek-mae-results}
\end{table}

\section{DeepSeek v4 Results}
\label{sec:appendix-deepseek}

We repeat our analysis on DeepSeek v4 models (\texttt{DeepSeek-v4-Flash} as the task LLM; \texttt{DeepSeek-V4-Pro} for loss, gradient, and optimizer LLMs). These provide a stronger starting point for the task-following and optimization steps. We use the MAE validation gate and run $N\!=\!3$ seeds.

\subsection{Overall Performance}
\label{sec:appendix-deepseek-main-results}

Table~\ref{tab:deepseek-mae-results} summarizes performance at step 12. We observe a stronger backbone for the task and optimization LLMs result in a superior increase from the initial prompt.
Multi-objective modes start from a stronger initialization ($\rho\approx0.38$) and reach best avg~$\rho$ between $0.409$ and $0.429$. \textsc{SSC} attains the best avg~$\rho$ and the highest HVI.
However, we observe that the trend sustains as before, with the \textit{improvement} through prompt optimization being highest for single-objective optimizations.

Figure~\ref{fig:deepseek-trajectories} plots the per-task optimization trajectories. Unlike the Qwen3 setting, where multi-objective modes stagnated at or below the initial prompt (Figure~\ref{fig:trajectories}), the DeepSeek backbone yields consistent improvement from initialization. 

The pattern of improvement, however, mirrors the main results: Single-Task achieves the largest gain across steps, while the multi-objective modes show flatter trajectories. The stronger optimizer LLM lifts all modes substantially, but it does not close the gap between single-objective and multi-objective optimization. 

Per-task traces remain bounded within narrower bands than their Qwen3 counterparts, consistent with the observation that a stronger task model produces more stable evaluation dynamics.

\begin{figure}[h!]
\centering
\includegraphics[width=0.95\linewidth]{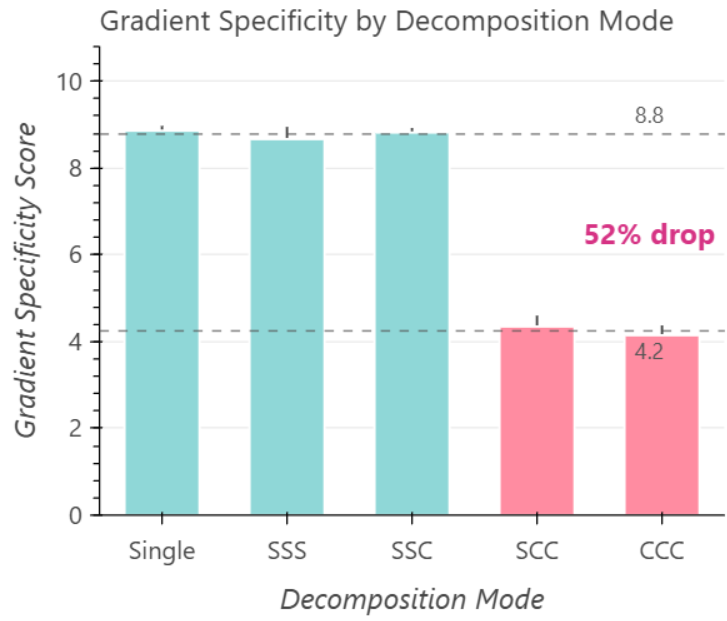}
\caption{Gradient specificity by decomposition mode under DeepSeek.}
\label{fig:deepseek-specificity}
\end{figure}

\subsection{Gradient Specificity}
\label{sec:deepseek-specificity}
With DeepSeek, the gradient dilution cliff persists: per-task gradient modes (Single/SSS/SSC) remain highly task-focused (mean $8.78$ specificity), while joint-gradient modes (SCC/CCC) drop to mean $4.25$ (a $52\%$ reduction), indicating that combining criteria in a single gradient call yields substantially more generic feedback.
As shown in Figure~\ref{fig:deepseek-specificity}, gradient specificity remains high for per-task gradient modes, while joint-gradient modes exhibit a sharp drop under the DeepSeek configuration.

\section{Multi-Objective Judge Prompt Template}
\label{sec:appendix-prompts}

Our multi-criteria prompt has two parts: a \textbf{frozen skeleton} (evaluation directive, output format) and \textbf{mutable per-task instructions} (the ``Instructions'' section). 

Only the per-task instructions are updated during optimization; the skeleton remains fixed throughout all 12 steps. Our skeleton is specific to \SummEval{} but may be easily adapted to other problems.

{
\begin{prompt}[title={Initial prompt template}]
\scriptsize
\begin{verbatim}
You are a careful, calibrated evaluator. Your goal is to
produce an accurate evaluation by following the 
Instructions below.

## Task
Evaluate the Summary given the Source Text using the
Instructions below.
1. Consider every strength and flaw you find when making 
your evaluation.
2. Based on the number and severity of the strengths and
flaws, assign a value.
Use the Instructions below to perform your evaluation. 
Output a JSON with the requested scores. Do NOT include
reasoning or explanations.

## Output format (follow this EXACTLY):
{
  "fluency": 1|2|3|4|5,
  "relevance": 1|2|3|4|5,
  "coherence": 1|2|3|4|5,
  "consistency": 1|2|3|4|5
}

## Instructions:          
- fluency: Rate from 1 to 5.
- relevance: Rate from 1 to 5.
- coherence: Rate from 1 to 5.
- consistency: Rate from 1 to 5.


## Sample:
Summary: <summary text>
Source Text: <source text>
\end{verbatim}
\end{prompt}
}

After optimization, the per-task instructions expand substantially. Below is an example from a SSS run (MAE validation, step 1), showing how the optimizer rewrites the generic one-line instructions into detailed, criterion-specific rubrics.

{
\begin{prompt}[title={Example optimized instructions (step 1)}]
\scriptsize
\begin{verbatim}
## Instructions:
- fluency: Fluency: Rate 1-5 based on
  grammatical correctness, sentence flow,
  and ease of parsing. A score of 5 reflects
  natural, effortless readability even with
  minor syntactic informality. Do not penalize
  for stylistic choices that do not impede
  comprehension.
- relevance: Relevance: Rate 1-5 based on how
  completely the summary addresses the core
  informational needs implied by the source
  text. A score of 5 means all key facts,
  outcomes, and implications are included with
  no deviation. Relevance penalizes both
  omission (missing key facts) and commission
  (adding unsupported claims).
- coherence: Coherence: Rate 1-5 based on
  logical flow, temporal sequence, and
  referential stability. A score of 5 means
  events unfold in a cause-effect or
  chronological order, with clear antecedents
  for all pronouns and noun phrases.
- consistency: Consistency: Rate 1-5 based on
  whether the summary contradicts any claim
  in the source text. A score of 5 requires
  no factual or inferential contradictions.
  A single major inconsistency reduces the
  score to 1.
\end{verbatim}
\end{prompt}
}

\section{Diagnostic Evaluation Prompts}
\label{sec:diagnostic-prompts}

Below we include the prompts used for task-level diagnostics. Both diagnostics are evaluated post-hoc by Claude Sonnet~4.6.

\subsection{Gradient Specificity Evaluator}
The following prompt rates each textual gradient on a 1 to 10 scale for task-specificity:

{
\begin{prompt}[title={Gradient specificity evaluator prompt}]
\scriptsize
\begin{verbatim}
You are measuring how focused a piece of
textual feedback (called a "gradient") is on
a specific evaluation task, versus being
diluted with generic advice or advice that
belongs to other tasks.

The target task is "{task}". The possible
tasks are: fluency, relevance, coherence,
consistency.

Rate from 1 to 10 how well this gradient
focuses on the "{task}" task.
1 = completely generic or mostly addresses
    other tasks.
10 = laser-focused on "{task}" with concrete,
     task-specific fixes.

## The Gradient
{gradient_text}

Respond with ONLY a single integer from 1
to 10. No explanation.
\end{verbatim}
\end{prompt}
}

\subsection{Feedback Adherence Evaluator}
The following prompt measures how well the optimizer incorporated each gradient into its instruction edit:

{
\begin{prompt}[title={Feedback adherence evaluator prompt}]
\scriptsize
\begin{verbatim}
You are evaluating whether revisions to
task-specific instructions correctly
addressed the gradient (suggested changes).

The instructions are for an LLM judge that
evaluates the "{task}" task. The Gradient
may contain suggestions about multiple
tasks; consider only suggestions pertaining
to "{task}".

Rate from 1 to 10 how well the New
Instructions address the Gradient for
"{task}".
1 = completely ignores/contradicts.
10 = precisely addresses every point while
     preserving what worked.

## Old Instructions
{old_instruction}

## New Instructions
{new_instruction}

## Gradient (Suggested Changes)
{gradient_text}

Respond with ONLY a single integer from 1
to 10. No explanation.
\end{verbatim}
\end{prompt}
}

\end{document}